\pdfoutput=1
\documentclass[11pt]{article}

\usepackage[]{acl}

\usepackage{times}
\usepackage{xspace}
\usepackage[T1]{fontenc}
\usepackage[utf8]{inputenc}

\usepackage{microtype}

\usepackage{cite}
\usepackage{algorithmic}
\usepackage{graphicx}
\usepackage[cjk]{kotex}
\usepackage{latexsym}
\usepackage[ruled,vlined,linesnumbered]{algorithm2e}
\usepackage{lipsum}
\usepackage{amsmath,amssymb,amsfonts}
\usepackage{caption}
\usepackage{subcaption}
\usepackage{textcomp}
\usepackage{url}
\usepackage{enumitem}
\usepackage{mathtools}
\usepackage{cuted}
\usepackage{nccmath}
\usepackage{makecell}
\usepackage{multirow}
\usepackage{booktabs}
\usepackage{array}
\usepackage{todonotes}
\usepackage{natbib}

\title{QUAK: A Synthetic Quality Estimation Dataset for Korean-English \\ Neural Machine Translation}

\author{Sugyeong Eo$^1$, Chanjun Park$^{1,2}$, Hyeonseok Moon$^1$, Jaehyung Seo$^1$, \\ \bf \large Gyeongmin Kim$^1$, Jungseob Lee$^1$, Heuiseok Lim$^1$\thanks{$^*$ Corresponding Author} \\
$^1$Korea University, $^2$Upstage \\
\small{\texttt{{\{djtnrud,bcj1210,glee889,seojae777,totoro4007,omanma1928,limhseok\}@korea.ac.kr}}} \\ \small{\texttt{chanjun.park@upstage.ai}}}

\begin{document}
\maketitle
\begin{abstract}
With the recent advance in neural machine translation demonstrating its importance, research on quality estimation (QE) has been steadily progressing. QE aims to automatically predict the quality of machine translation (MT) output without reference sentences. Despite its high utility in the real world, there remain several limitations concerning manual QE data creation: inevitably incurred non-trivial costs due to the need for translation experts, and issues with data scaling and language expansion. To tackle these limitations, we present QUAK, a Korean-English synthetic QE dataset generated in a fully automatic manner. This consists of three sub-QUAK datasets QUAK-M, QUAK-P, and QUAK-H, produced through three strategies that are relatively free from language constraints. Since each strategy requires no human effort, which facilitates scalability, we scale our data up to 1.58M for QUAK-P, H and 6.58M for QUAK-M. As an experiment, we quantitatively analyze word-level QE results in various ways while performing statistical analysis. Moreover, we show that datasets scaled in an efficient way also contribute to performance improvements by observing meaningful performance gains in QUAK-M, P when adding data up to 1.58M.
\end{abstract}

\section{Introduction}
Quality estimation (QE) is the task of predicting the translation quality as a continuous value or discrete tags by referring to a source sentence and its machine translation (MT) output \citep{blatz2004confidence,specia2009estimating,specia-etal-2013-quest}. 
Since quality annotations on MT output are applied in various ways according to the granularity levels (word, sentence, document, etc.), QE research has been constantly developing in recent years \citep{kim2017predictor,fomicheva2020multi,alva2021deepquest,ding2021levenshtein}.

\begin{table}
\resizebox{0.5\textwidth}{!}{%
\begin{tabular}{@{}c|l@{}}
\toprule[1.5pt]
\textbf{MT output}& Given that the Chinese authorities do not deny it , \textcolor{purple}{it is highly likely} .\\
\textbf{pseudo-PE}& Given that the Chinese authorities do not deny it , \textcolor{purple}{chances are high} . \\ 
\multirow{2}{*}{\textbf{MT output tags}}&OK OK OK OK OK OK OK OK OK OK OK OK OK OK OK OK\\
& OK OK OK OK OK \textcolor{purple}{BAD} OK \textcolor{purple}{BAD} OK \textcolor{purple}{BAD} OK \textcolor{purple}{BAD} OK OK OK \\ \midrule
\textbf{Source}& 중국 당국이 부인하지 않는 것으로 볼 때 \textcolor{teal}{가능성이 높다} . \\
\textbf{Source tags}& OK OK OK OK OK OK OK \textcolor{teal}{BAD BAD} OK \\ 
\textbf{\makecell{Alignments}} & 0-3 1-4 2-7 3-5 3-6 4-8 5-8 6-0 \textcolor{teal}{7-13 8-11 8-12} 9-14\\ \midrule
\multirow{2}{*}{\textbf{Edits}}& \textcolor{black}{(1) Insertion (` '$\to$ it) (2) Substitution (chances$\to$is)}\\
&\textcolor{black}{(3) Substitution (are$\to$highly) (4) Substitution (high$\to$likely)} \\ \bottomrule[1.5pt]
\end{tabular}}\caption{\label{tb:example}An example of QUAK dataset. For the correct translation, one insertion and three substitutions are required for the MT output. Although not included in this example, if there is a missing word, a BAD tag is attached to the location of the corresponding gap token. We indicate the alignment information (Alignments) in the form of \{source index\}-\{aligned MT output index\}.}
\end{table}

Owing to this importance, datasets for training QE systems are being released continuously. However, we highlight three limitations for the existing QE dataset. (1) First, non-trivial human labor and time cost are required when constructing data. Source sentences, MT output, and quality annotations are dataset prerequisites for QE learning, among which translation experts proficient in a language pair are essential in the labeling process. Employing experts is far more difficult especially in low-resource languages.

(2) As an extension of the first limitation, manual QE datasets are restricted in size regardless of the data resource. The meticulous work of creating human post-edited sentences with minimal modifications slows down the construction time, which makes it difficult to scale. Most released QE datasets, including those from the Conference on Machine Translation (WMT) are composed of data less than 10K in size \citep{fujita2017japanese,fomicheva2020mlqe}. This is a much rarer amount compared with the large volumes of data used by studies on GPT 3 \citep{brown2020language} in terms of data-hungry NLP. 

(3) Available QE language pairs are limited. Although the released WMT QE dataset considers high, medium, and low resources \citep{fomicheva2020unsupervised}, it still covers only a much smaller number of language pairs compared with parallel corpora. Since data construction in opposite directions for a language pair requires an entirely different human post-edited sentence, numerous language pairs and directions are yet to be utilized.

To mitigate the above limitations, we introduce QUAK \footnote{Our QUAK dataset is publicly available at \texttt{\textcolor{blue}{\url{https://www.aihub.or.kr/aihubdata/data/view.do?currMenu=120&topMenu=100&aihubDataSe=extrldata&dataSetSn=71268}}}.}, a large-scale Korean-English synthetic QE dataset. This is built as an automated process by taking the \newcite{eo-etal-2021-dealing} approach and aims to train word-level QE. Namely, the data generation process does not demand human post-editing, allowing data to be built at scale than manual methods. In addition, language extension is relatively free as it is language-agnostic within a language pair in which Google translation is possible and a corresponding corpus exists. Therefore, we adopt Korean-English, one of the morphologically rich languages rarely addressed in the QE field.

For constructing QUAK, A monolingual or parallel corpus and an MT model are required. QUAK is divided into three sub-QUAK datasets according to data sources: (1) QUAK-Monolingual (QUAK-M) leveraging a monolingual corpus of the target language, (2) QUAK-Parallel (QUAK-P) leveraging a parallel corpus, and (3) QUAK-Hybrid (QUAK-H) jointly leveraging monolingual and parallel corpus. The final QUAK training data size in QUAK-P and QUAK-H is 1.58M and 6.58M in QUAK-M, which is about 225 times and 940 times larger than the 7K size of the WMT official dataset \citep{fomicheva2020mlqe}.

Considering that QUAK is synthetic data, we scrutinize the dataset with statistics and a quantitative analysis to provide reliability and quality assurance. In the quantitative analysis, in particular, we first compare the word-level QE model fine-tuning performance based on multiple multilingual pre-trained language models (mPLMs) using only 100K pieces from each sub-QUAK. Thereafter, we use the best performing model to incrementally scale the data size and track performance fluctuations.

As a result of the experiment, the XLM-RoBERTa (XLM-R) \citep{conneau2019unsupervised} large model is the most competitive, showing a difference of up to 0.12 MCC compared to other mPLMs such as multilingual BART (mBART) \citep{liu2020multilingual}, XLM \citep{lample2019cross}. Furthermore, scaling the data to 1.58M tends to improve overall model performance. Based on the MT output-side, QUAK-M obtains performance gain of maximum 0.042, QUAK-P 0.037, and QUAK-H 0.029 MCC. Our contributions are as follows:

\begin{itemize}
    \item To minimize the exorbitant human-demand and time cost of QE, we construct and release the QUAK dataset in a fully automatic manner, exploiting three efficient data generation strategies.
    
    \item To address the size limitation, we expand the synthetic data to a large-scale. We scale QUAK up to 940 times compared with the WMT official dataset.
    
    \item As the language pair for QUAK, we choose Korean-English, a low-resource language pair that has never been released before. Language coverages can be extended if only the translation model and its corpus are satisfied in the generation process.
    
    \item We analyze the QE fine-tuning performance according to various mPLMs, and analyze the results of progressively expanding the data for the best performing QE model.
    
\end{itemize}

\begin{figure*}[h]
 \includegraphics[width=1.0\linewidth, height=3.8cm]{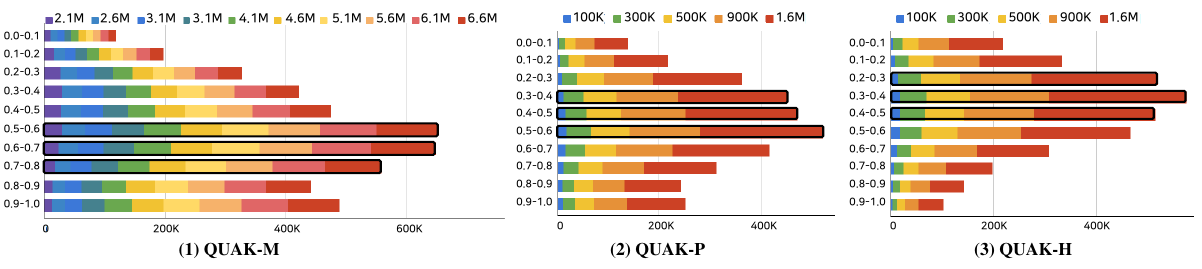}
 \caption{Distribution of the data size according to the TER range for each sub-QUAK. We present the top three scopes with the largest amount in bold lines.}
 \label{fig1:data_stat}
\end{figure*}

\section{Related Work}\label{sec:related_work}
As QE research has increasingly been introduced recently, human-labeled QE datasets are also being released \citep{specia2010dataset,fujita2017japanese,fomicheva2020unsupervised,fomicheva2020mlqe}. However, data construction processes have several limitations in terms of time cost, data size, and available language pairs. 

Many studies have been conducted continuously to handle these limitations. To name a few, \newcite{tuan2021quality} propose a synthetic data construction method that utilizes a parallel corpus to alleviate the cost of human labor and time cost. In such study, translation errors committed through a language model or an NMT system are injected to parallel sentences. A similar method of generating synthetic data through parallel corpus has also been leveraged in automatic post-editing research \citep{negri2018escape}.

To address data size restrictions caused by time cost in human annotations, attempts have been made using data augmentation \citep{lee2020two,wang2020hw,gajbhiye2021knowledge,ding2021jhumicrosoft} and unsupervised learning \citep{etchegoyhen2018supervised}. In the study of \newcite{fomicheva2020unsupervised}, unsupervised quality indicators based on uncertainty quantification are exploited to train the QE model.

To tackle constraints about available language pairs, cross-lingual zero-shot QE approaches are constantly studied \citep{sun2020exploratory,eo2021study}. Following this trend, WMT21 includes zero-shot to their main interests, evaluating the competence of a QE model on unseen languages \citep{specia2021findings}. This study addresses all of the respective factors and presents QUAK.

\section{QUAK}
QUAK is a QE dataset for evaluating the quality of Korean-English MT output, and includes three sub-QUAKs according to the data selection. Each sub-QUAK comprises \textit{(1) a source sentence, (2) its MT output, and (3) OK/BAD quality annotation}. Source sentence and its MT output are utilized as model input, allowing the model to classify the quality of these sentences into OK/BAD tags on a token basis. Quality annotations separately exists for source sentence and MT output. When the model predicts the translation quality, these are used as a ground-truth for evaluation. 
With primary consideration of efficiency and effectiveness, we construct data in a fully-automatic manner, exploited method by \newcite{eo-etal-2021-dealing}. We select existing monolingual or parallel corpora for our data sources (detailed in Section \ref{subsec:data_sources}) and use them to each data production process (detailed in Section \ref{subsec:const_process}).

\subsection{Dataset Sources} \label{subsec:data_sources}
QUAK is divided into three sub-QUAKs. The raw dataset requirements to build each sub-QUAK are as follows: QUAK-M requires a target language monolingual corpus, QUAK-P requires a parallel corpus, QUAK-H requires both corpora. 

For a monolingual corpus, we adopt English Wikipedia, which consists of documents on a wide range of topics. We use it to handle the various translation errors that the MT model may commit to the diverse entities and expressions in Wikipedia. We randomly extract 5M sentences to generate QUAK-M. For a parallel corpus, we leverage AI hub parallel corpus released by Korea National Information Society Agency\footnote{\textcolor{blue}{\url{https://aihub.or.kr/}}}. AI hub corpus also covers various fields such as news, journals, law, and culture, and is produced with high quality through human inspection. The parallel corpus contains 1,602,002 pairs.

For fair comparison with other sub-QUAKs, in the case of QUAK-M, we combine both AI hub and Wikipedia source. Namely, we configure 1.58M of QUAK-M using the target-side text of the Ai hub and the remaining 5M using Wikipedia. The validation and test set is configured by randomly selecting 12K pieces of AI hub data.

\subsection{Dataset Construction Process} \label{subsec:const_process}
\paragraph{QUAK-M} 
For QUAK-M, we utilize a target language monolingual corpus. With the text, we first conduct a round-trip translation. We translate the English corpus into Korean sentences, where we denote them as pseudo-source sentences. We once again forward-translate the pseudo-source to generate the MT output. 

When pseudo-source and its MT output have been created, quality annotations are tagged. Prior to label annotation, we pre-define target language monolingual corpus to be a flawless sentence. Based on this assumption, we consider this text as a pseudo-post-edited (pseudo-PE) sentences for which correction has been completed. By comparing the MT output and pseudo-PE in a token-wise fashion, we measure the minimum substitution/deletion/insertion errors needed based on edit distance. The OK/BAD tag indicating correct/wrong translation for each token in the MT output is further annotated. If the number of MT output tokens in a sentence is $N$, the number of OK/BAD tags for this is $2N +1$ because gap tokens are attached to the front and back of each MT output token. If there are missing words, BAD tags are added to the position of the corresponding gap token, otherwise OK tags are labeled. Apart from tagging the MT outputs, source tags are also annotated according to the binary tags of the MT outputs based on word alignment information. The tag for the source sentence excludes the gap token labeling.

\paragraph{QUAK-P} 
The parallel corpus is leveraged in the QUAK-P configuration. This has higher connectivity between the source and target sides and has an intact source sentence compared with the pseudo-source of QUAK-M. To obtain QUAK-P, we proceed a one-way translation from the source to target language. In this case, source-side difference from QUAK-M leads to various translation results for the MT output. Similar to the QUAK-M generation process, we consider the target-side text of the parallel corpus as a pseudo-PE. With source sentences, its MT output, and pseudo-PE, we label the quality of the translation results. After calculating the minimum edit operation between the MT output and the pseudo-PE, BAD tags are attached to the token where the modification occurred. For quality annotations on source sentences, the same tags are attached to the MT output index and the aligned source index.

\paragraph{QUAK-H} 
In QUAK-H, we combine the above two previous sub-QUAKs to generate various translation results with a limited corpus. We compose the source sentence and MT output by selectively utilizing two approaches proposed in QUAK-P and QUAK-M, respectively. Namely, we use the source-side text and pseudo-PE text from QUAK-P, and the MT output-side text from QUAK-M. By dealing with two different MT outputs with the same source-side text in QUAK-P, we induce the QE model to learn by referring to various combinations of the source sentence and MT output. For the next step, we tag labels for quality annotation as mentioned above.

\paragraph{Final constructed QUAK dataset} \label{subsec:example}
After three construction processes, we obtain a total of 1,578,002 training examples for QUAK-P and QUAK-H, and 6,578,002 training examples for QUAK-M. We present an example of QUAK in Table~\ref{tb:example}. MT output for a source sentence ``중국 당국이 부인하지 않는 것으로 볼 때 가능성이 높다 .'' is mistranslated into ``\texttt{Given that the Chinese authorities do not deny it, it is highly likely .}''. The MT output should be corrected into ``\texttt{chances are high}''. This should perform a four minimum correction, which will result in a four BAD tags of the entire MT output tag. In addition, based on the word alignment ``가능성(7)--\texttt{likely}(13), 높다(8)--\texttt{is}(11), 높다(8)--\texttt{highly}(12)'', the BAD tag index of the MT output is also reflected in the source-side index.

\section{Experimental Setup}
\subsection{Experimental Design}
In this section, we present the statistical and quantitative analysis done on the QUAK. In the statistical analysis, we measure the sentence length, token length, and average token length per sentence for each sub-QUAK. We also calculate the mean, median, standard deviation, and variance of the translation edit rate (TER) score. Regarding tags, we count the total number of OK and BAD tags. 

During the quantitative analysis, we experiment three word-level QE fine-tuning to efficiently achieve high performance and analyze large-scale QUAK data. In the first experiment, we fine-tune multiple mPLMs with 100K pieces of QUAK-M, P, H to explore which model performs better for QUAK. 

Thereafter, we inspect the impact on the amount of QUAK. We fine-tune the data for the previous best performing model, scaling each sub-QUAK exponentially from 100K to 1.58M. As mentioned earlier, one consideration is that QUAK-M (1.58M) consists of target-side text in a parallel corpus for proper comparison with the data generated by other strategies.

In the last experiment, we gradually increase the size of QUAK-M. Our result includes the corresponding performance while extending from the previous size of 1.58M to 6.58M in 500K increments.

\begin{table}[]
\resizebox{0.48\textwidth}{!}{
\begin{tabular}{c|cccc}
\toprule[1.5pt]
\multicolumn{1}{c|}{\textbf{Attributes}}  & \textbf{Google} & \textbf{Amazon} & \textbf{Microsoft} & \textbf{Systran} \\ \midrule[1.5pt]
\# of Source Sentences& 12,000 & 12,000 & 12,000 & 12,000\\
\# of MT Output& 12,000 & 12,000 & 12,000 & 12,000\\
\# of pseudo-PE    & 12,000 & 12,000 & 12,000 & 12,000\\ \midrule
\# of Source Tokens& 199,413& 199,413& 199,413& 199,413 \\
\# of MT Output Tokens& 340,264& 303,535& 325,973& 346,030 \\
\# of pseudo-PE Tokens    & 342,385& 342,385& 342,385& 342,385 \\ \midrule
Average Token Per Source Sentence& 16.62& 16.62& 16.62  & 16.62\\
Average Token Per MT Output& 28.36& 25.29& 27.16  & 28.84\\
Average Token Per pseudo-PE    & 28.53& 28.53& 28.53  & 28.53\\ \midrule
Mean TER & 0.57& 0.63& 0.63   & 0.46 \\
Median TER& 0.57& 0.64& 0.64   & 0.44 \\
STD TER  & 0.23& 0.21& 0.21   & 0.26 \\
Variance TER & 0.05& 0.04& 0.05   & 0.07 \\ \midrule
\# Source OK tags& 112,647& 94,562 & 97,825 & 134,503 \\
\# Source BAD tags    & 86,766 & 104,851& 101,588& 64,910\\
\# MT Output OK tags  & 510,085& 429,775& 465,441& 560,221 \\
\# MT Output BAD tags & 182,443& 189,295& 198,505& 143,839 \\ \bottomrule[1.5pt]
\end{tabular}}\caption{\label{tb:test}Statistics for the four test sets. We denote the target-side text of corpus as pseudo-PE.}
\end{table}

\begin{table*}
\centering
\resizebox{0.8\textwidth}{!}{%
\begin{tabular}{c|ccc|ccc} 
\toprule[1.5pt]
\multicolumn{1}{l|}{}& \multicolumn{3}{c|}{\textbf{Train}}   & \multicolumn{3}{c}{\textbf{Valid}}  \\ \midrule
\textbf{Attributes} & \textbf{QUAK-M}& \textbf{QUAK-P}     & \textbf{QUAK-H}     & \textbf{QUAK-M}  & \textbf{QUAK-P}  & \textbf{QUAK-H} \\ 
\midrule[1.5pt]
\# of Source Sentences & 6,578,002   & 1,578,002  & 1,578,002  & 12,000  & 12,000  & 12,000 \\
\# of MT Output & 6,578,002   & 1,578,002  & 1,578,002  & 12,000  & 12,000  & 12,000 \\
\# of pseudo-PE & 6,578,002   & 1,578,002  & 1,578,002  & 12,000  & 12,000  & 12,000 \\ 
\midrule
\# of Source Tokens & 92,848,776  & 25,149,673 & 25,149,673 & 209,894 & 199,624 & 199,624\\
\# of MT Output Tokens & 139,620,328 & 42,051,001 & 39,850,492 & 318,959 & 340,855 & 318,921\\
\# of pseudo-PE Tokens     & 148,922,086 & 42,103,966 & 42,103,966 & 342,021 & 341,996 & 341,996\\ 
\midrule
Average Token Per Source Sentence & 14.12 & 15.94& 15.94& 17.50   & 16.64   & 16.64  \\
Average Token Per MT Output & 21.23 & 26.65& 25.25& 26.58   & 28.40   & 26.58  \\
Average Token Per pseudo-PE & 22.64 & 26.68& 26.68& 28.50   & 28.50   & 28.50  \\ 
\midrule
Mean TER   & 0.61& 0.50 & 0.42 & 0.45    & 0.57    & 0.45\\
Median TER & 0.63& 0.50 & 0.40 & 0.44    & 0.56    & 0.44\\
STD TER   & 0.24& 0.25 & 0.23 & 0.21    & 0.23    & 0.21\\
Variance TER & 0.06& 0.06 & 0.05 & 0.04    & 0.05    & 0.04\\ 
\midrule
\# Source OK tags   & 50,421,860  & 15,600,416 & 16,269,784 & 84,873  & 113,192 & 121,700\\
\# Source BAD tags  & 42,426,916  & 9,549,257  & 8,879,889  & 82,343  & 86,432  & 77,924 \\
\# MT Output OK tags& 204,721,896 & 65,852,185 & 65,033,087 & 339,304 & 511,542 & 506,437\\
\# MT Output BAD tags  & 81,096,762  & 19,827,819 & 16,245,899 & 157,308 & 182,168 & 143,405\\
\bottomrule[1.5pt]
\end{tabular}
}
\caption{\label{tb:stat}Statistics for three sub-QUAK training and validation set}
\end{table*}

\subsection{Experimental Settings}
\paragraph{Models}
In all experiments, we exploit the MicroTransQuest \citep{ranasinghe2021exploratory} framework. While it only uses an XLM-R model, we utilize additional mPLMs: In the QE model training, we leverage XLM, XLM-R, and mBART. 

From Huggingface \citep{wolf2019huggingface}, we load five mPLMs that have learned both Korean and English: xlm-mlm-100-1280, xlm-roberta-base, xlm-roberta-large, facebook/mbart-large-cc25, and facebook/mbart-large-50.

\paragraph{Datasets}
For the data construction, the following tools are used in this study. As monolingual data we dump Wikipedia and use Wikiextractor\footnote{\textcolor{blue}{\url{https://github.com/attardi/wikiextractor}}} to extract plain text. We train a Korean-English and English-Korean MT model using the fairseq \citep{ott2019fairseq} package with SentencePiece subword tokenization \citep{kudo2018sentencepiece} to translate sentences. We train the word alignment between the source and target text using the FastAlign \citep{dyer2013simple} toolkit and measure the edit distance using Tercom software \citep{snover2006study}. Tag annotation is executed using the Unbabel corpus builder\footnote{\textcolor{blue}{\url{https://github.com/Unbabel/word-level-qe-corpus-builder}}}. We adopt Mosesdecoder \citep{koehn2007moses} for additional data preprocessing.

\paragraph{Evaluation}
For constructing the test sets, we utilize a publicly available external machine translator to ensure the reliability and objectivity of the QE results. Four representative commercialized machine translators are adopted, including Google\footnote{\textcolor{blue}{\url{https://translate.google.co.kr/?hl=en}}}, Amazon\footnote{\textcolor {blue}{\url{https://aws.amazon.com/translate/}}}, Microsoft\footnote{\textcolor{blue}{\url{https://www.microsoft.com/en-us/ translator/}}}, and Systran\footnote{\textcolor{blue}{\url{https://translate.systran.net/}}}. Through these, the test sets are established in the same manner as the strategy used in QUAK-P. The test sets are based on 12K sentence pairs randomly extracted from the Ai hub parallel corpus without overlapping with the training and validation sets. Matthews correlation coefficient (MCC) \citep{chicco2020advantages} is used as a metric for evaluating QE model performance.

Table \ref{tb:test} provides the test set statistics. When analyzing the number of OK/BAD tags, the results vary depending on the translator even when the same source sentence is used. Systran differs the most compared with other test sets: there are 143,839 MT output BAD tags with an average TER difference of 0.17 with the highest value of 0.63.

\section{Analysis and Results}
\subsection{Data statistics and analysis}\label{sec:stat}
We report the statistics for QUAK in Table \ref{tb:stat}. QUAK-M additionally uses the English Wikipedia corpus consisting of 5M samples, and this yields different data sizes in contrast to other sub-QUAKs. Comparing training set of QUAK-M with other sub-QUAKs, the most dominant part is the relation between the average token length and TER score. QUAK-P and QUAK-H show lower TER scores even though their average tokens per sentence are relatively higher. We interpret this result as a case where the translation works well even if the average sentence length is long. As shown in Figure \ref{fig1:data_stat}, comparing data sizes by TER range is also consistent with this statistic. The data is mainly concentrated in the 0.5--0.8 range for QUAK-M, 0.3--0.6 for QUAK-P, and 0.2--0.5 for QUAK-H. From this, we speculate that Wikipedia may have more noise in the text itself than the Ai hub, and that a large number of errors are committed during the translation process.

Next, for QUAK-P and QUAK-H, the number of BAD tags of QUAK-P is greater than that of QUAK-H in the MT output. It is noteworthy that the MT output of QUAK-H is created based on a round-trip translation, which is identical to that of QUAK-M. These indicate that the pseudo-source generated by the target language text of Ai hub is adequately restored to the original sentence when translated back, even if it is different from the correctly translated source-side of the parallel corpus.

\subsection{Experimental Results} 
\paragraph{Performance Comparison by mPLMs} \label{sec:result_mplms}
We provide the fine-tuning results for mPLMs by selecting only 100K of the datasets in Table \ref{tb:model}. From the experimental results, XLM-R-large model shows the best performance. Based on the MT output-side MCC (Target MCC) of the Google test set, XLM-R-large reports 0.366, 0.401, and 0.324 for QUAK-M,P,H, and outperforms the XLM-R-base model by 0.023, 0.024, and 0.004, respectively. For the source-side MCC (Source MCC), XLM-R-large also achieves 0.285, 0.331, and 0.271 for QUAK-M,P,H, which are the best competencies compared to other models.

\begin{figure*}[h]
 \centering
 \includegraphics[width=1.0\linewidth]{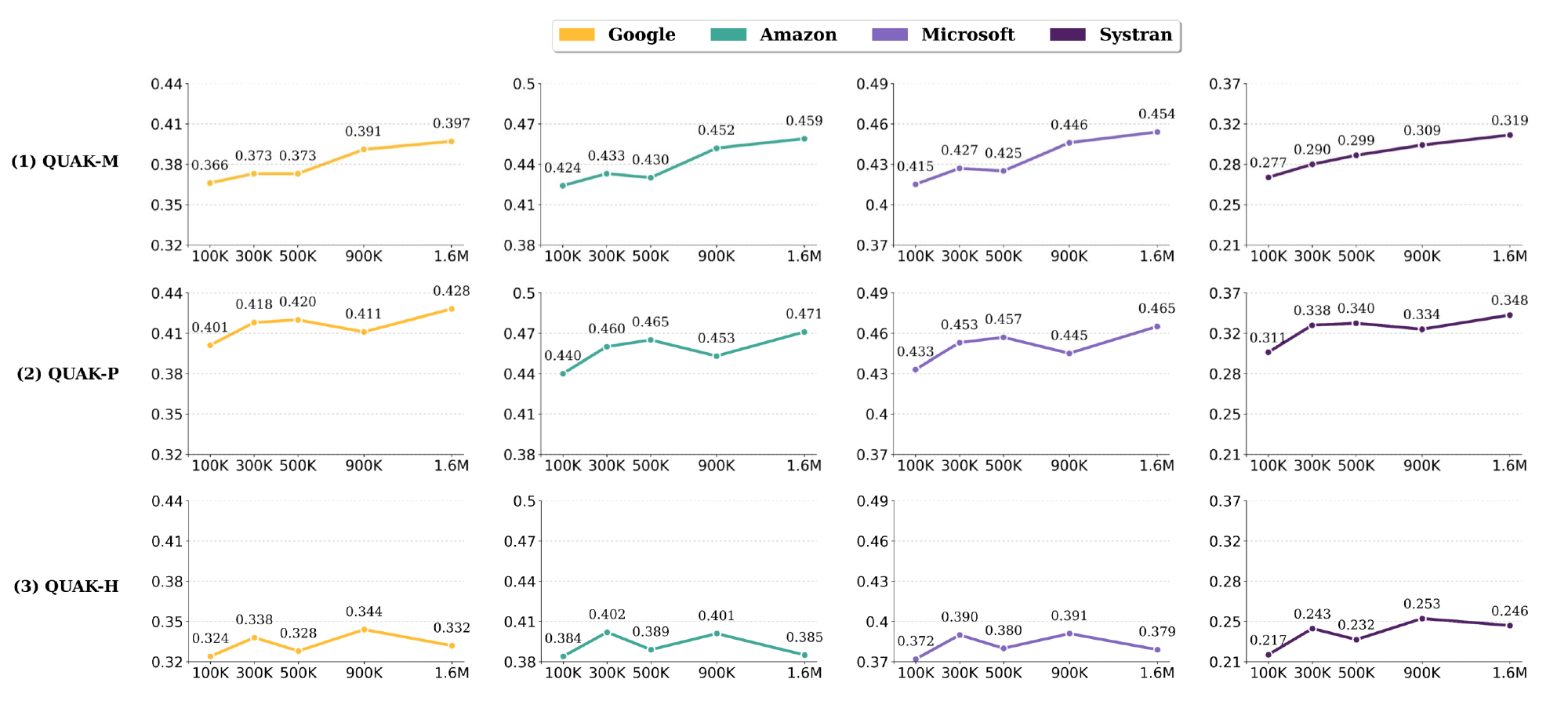}
 \caption{\label{fig2:result_grad} MCC variation of QUAK according to data scaling
 }
\end{figure*}

\begin{table}
\centering
\resizebox{0.48\textwidth}{!}{
\begin{tabular}{c|c|ccccc} 
\toprule[1.5pt]
& \textbf{Dataset} & \begin{tabular}[c]{@{}l@{}}\textbf{XLM-R}\\\textbf{-base}\end{tabular} & \begin{tabular}[c]{@{}l@{}}\textbf{XLM-R}\\\textbf{-large}\end{tabular} & \textbf{mBART} & \textbf{mBART50} & \textbf{XLM} \\ 
\midrule[1.5pt]
\multicolumn{7}{c}{\textbf{Google}}    \\ 
\midrule
\multirow{3}{*}{Target MCC} & QUAK-M& 0.343  & \textbf{0.366}   & 0.340 & 0.343& 0.296\\
& QUAK-P& 0.377  & \textbf{0.401}   & 0.376 & 0.382& 0.339\\
& QUAK-H& 0.320  & \textbf{0.324}   & 0.306 & 0.314& 0.292\\ 
\midrule
\multirow{3}{*}{Source MCC} & QUAK-M& 0.279  & \textbf{0.285}   & 0.275 & 0.276& 0.231\\
& QUAK-P& 0.315  & \textbf{0.331}   & 0.309 & 0.320& 0.285\\
& QUAK-H& 0.266  & \textbf{0.271}   & 0.258 & 0.267& 0.249\\ 
\midrule[1.5pt]
\multicolumn{7}{c}{\textbf{Amazon}}    \\ 
\midrule
\multirow{3}{*}{Target MCC} & QUAK-M& 0.389  & \textbf{0.424}   & 0.388 & 0.385& 0.328\\
& QUAK-P& 0.408  & \textbf{0.440}   & 0.405 & 0.410& 0.362\\
& QUAK-H& 0.362  & \textbf{0.384}   & 0.264 & 0.359& 0.322\\ 
\midrule
\multirow{3}{*}{Source MCC} & QUAK-M& 0.324  & \textbf{0.342}   & 0.320 & 0.323& 0.254\\
& QUAK-P& 0.353  & \textbf{0.377}   & 0.341 & 0.354& 0.304\\
& QUAK-H& 0.305  & \textbf{0.323}   & 0.213 & 0.310& 0.276\\ 
\midrule[1.5pt]
\multicolumn{7}{c}{\textbf{Microsoft}}\\ 
\midrule
\multirow{3}{*}{Target MCC} & QUAK-M& 0.380  & \textbf{0.415}   & 0.382 & 0.380& 0.315\\
& QUAK-P& 0.401  & \textbf{0.433}   & 0.404 & 0.406& 0.346\\
& QUAK-H& 0.353  & \textbf{0.372}   & 0.253 & 0.355& 0.316\\ 
\midrule
\multirow{3}{*}{Source MCC} & QUAK-M& 0.307  & \textbf{0.329}   & 0.303 & 0.307& 0.244\\
& QUAK-P& 0.338  & \textbf{0.363}   & 0.327 & 0.338& 0.287\\
& QUAK-H& 0.290  & \textbf{0.310}   & 0.193 & 0.299& 0.271\\ 
\midrule[1.5pt]
\multicolumn{7}{c}{\textbf{Systran}}   \\ 
\midrule
\multirow{3}{*}{Target MCC} & QUAK-M& 0.261  & \textbf{0.277}   & 0.255 & 0.253& 0.206\\
& QUAK-P& 0.298  & \textbf{0.311}   & 0.289 & 0.296& 0.261\\
& QUAK-H& \textbf{0.226}  & 0.217   & 0.122 & 0.221& 0.196\\ 
\midrule
\multirow{3}{*}{Source MCC} & QUAK-M& \textbf{0.224}  & 0.223   & 0.218 & 0.221& 0.161\\
& QUAK-P& 0.247  & \textbf{0.250}   & 0.228 & 0.242& 0.210\\
& QUAK-H& \textbf{0.179}  & 0.174   & 0.076 & 0.176& 0.159\\
\bottomrule[1.5pt]
\end{tabular}}\caption{\label{tb:model}Comparison of word-level Korean-English QE performance by mPLMs fine-tuned with each sub-QUAK dataset}
\end{table}

In all test sets, except for Systran, the Target MCC and Source MCC of XLM-R-large performed the best in all sub-QUAK datasets. XLM-R-large differs from XLM-R-base in terms of the number of parameters; the former contains 550M, whereas the latter has 270M. Based on the Target MCC of the Amazon test set, XLM-R-large generally reports a higher performance than XLM-R-base, achieving 0.035, 0.032, and 0.022 higher values for QUAK-M, P, and H, respectively. These results demonstrate that the number of parameters in mPLMs poses a positive effect on the QE model learning.

Regarding mBART and mBART50, the latter outperforms the former in general. This implies the substantial impact of the number of pre-trained languages. It is noteworthy that mBART50 is pre-trained for 50 languages, enabling more multilingual support than mBART, which is learned on 25 languages. As Korean is regarded as a relatively low-resource language and especially utilizes only 100K data, we infer that mBART50 has more influence on competence gain from high-resource languages than mBART.

\paragraph{Performance Comparison for Scaling}

\begin{table}[]
\resizebox{0.48\textwidth}{!}{%
\begin{tabular}{l|lc|cc|cc|cc}
\toprule[1.5pt]
\multicolumn{1}{c|}{} & \multicolumn{2}{c|}{\textbf{Google}} & \multicolumn{2}{c|}{\textbf{Amazon}} & \multicolumn{2}{c|}{\textbf{Microsoft}} & \multicolumn{2}{c}{\textbf{Systran}} \\ \midrule

\begin{tabular}[c]{@{}c@{}}\textbf{Data}\\ \textbf{Size}\end{tabular} &  \begin{tabular}[c]{@{}c@{}}\textbf{Target}\\ \textbf{MCC}\end{tabular} & \begin{tabular}[c]{@{}c@{}}\textbf{Source}\\ \textbf{MCC}\end{tabular} & \begin{tabular}[c]{@{}c@{}}\textbf{Target}\\ \textbf{MCC}\end{tabular} & \begin{tabular}[c]{@{}c@{}}\textbf{Source}\\ \textbf{MCC}\end{tabular} & \begin{tabular}[c]{@{}c@{}}\textbf{Target}\\ \textbf{MCC}\end{tabular} & \begin{tabular}[c]{@{}c@{}}\textbf{Source}\\ \textbf{MCC}\end{tabular} & \begin{tabular}[c]{@{}c@{}}\textbf{Target}\\ \textbf{MCC}\end{tabular} &
\begin{tabular}[c]{@{}c@{}}\textbf{Source}\\ \textbf{MCC}\end{tabular} \\ \midrule[1.5pt]
1.58M & 0.397& 0.324& 0.459& 0.386& 0.454 & 0.376& 0.319& 0.273\\ \midrule
2.08M & 0.386& 0.319& 0.441& 0.371& 0.436 & 0.365& 0.317& 0.272\\ \midrule
2.58M & 0.384& 0.316& 0.442& 0.372& 0.436 & 0.364& 0.308& 0.264\\ \midrule
3.08M & 0.385& 0.316& 0.442& 0.375& 0.435 & 0.365& 0.313& 0.272\\ \midrule
3.58M & 0.381& 0.314& 0.443& 0.374& 0.437 & 0.365& 0.311& 0.269\\ \midrule
4.08M & 0.382& 0.313& 0.438& 0.369& 0.434 & 0.359& 0.310& 0.270\\ \midrule
4.58M  & 0.376& 0.310& 0.433& 0.368& 0.426 & 0.357& 0.307& 0.267\\ \midrule
5.08M & 0.378& 0.308& 0.436& 0.366& 0.432 & 0.362& 0.308& 0.269\\ \midrule
5.58M & 0.377& 0.319& 0.439& 0.374& 0.432 & 0.364& 0.307& 0.274\\ \midrule
6.08M & 0.351& 0.286& 0.400& 0.335& 0.398 & 0.334& 0.296& 0.261\\ \midrule
6.58M & 0.387& 0.325& 0.451& 0.379& 0.441 & 0.369& 0.313& 0.274\\ \bottomrule[1.5pt]
\end{tabular}%
}{\caption{\label{tb4:result_mono} Performance variation of QUAK-M according to data scaling}}
\end{table}

The previously obtained results show that the XLM-R-large model is superior to all 100K sub-QUAK datasets. For the next experiment, we explore the performance fluctuation by constantly increasing the size of the QUAK dataset to XLM-R-large. Figure \ref{fig2:result_grad} illustrates the variation of the performance depending on the corpus size. The experimental results demonstrate that the performance variation tends to be similar for all test sets.

When we exponentially scale the data for the three sub-QUAKs, QUAK-M had a notable achievement. Furthermore, QUAK-P showed a steady increase, except for the case of 900K. We confirm that data scaling is one factor in increasing the performance of the QE model. However, in the case of QUAK-H, there is no clear trend in terms of data expansion. We argue that although the MT output is applied owing to various translations for source sentences, the weakened connectivity between two sentences might impede learning.

\begin{table*}[]
\centering
\resizebox{0.70\textwidth}{!}{%
\begin{tabular}{c|c|c|c|c|c|c|c|c|c|c}
\toprule[1.5pt]
 & \multicolumn{10}{c}{\textbf{TER Range}} \\\midrule
\textbf{Data Size} & \textbf{0.0-0.1} & \textbf{0.1-0.2} & \textbf{0.2-0.3} & \textbf{0.3-0.4} & \textbf{0.4-0.5} & \textbf{0.5-0.6} & \textbf{0.6-0.7} & \textbf{0.7-0.8} & \textbf{0.8-0.9} & \textbf{0.9-1.0} \\\midrule[1.5pt]
\multicolumn{11}{c}{QUAK-P} \\\midrule
100K & 0.214 & 0.323 & 0.361 & 0.398 & 0.410 & 0.382 & 0.374 & 0.338 & 0.308 & 0.262 \\\midrule
1.58M & 0.209 & 0.348 & 0.386 & 0.420 & 0.438 & 0.413 & 0.395 & 0.377 & 0.325 & 0.279 \\\midrule
Diff & \underline{-0.005}& 0.025& 0.025& 0.022& \textbf{0.028}& \textbf{0.031}& 0.021& \textbf{0.039}& \underline{0.017}& \underline{0.017}\\\midrule[1.5pt]
\multicolumn{11}{c}{QUAK-H}    \\\midrule
100K & 0.206& 0.312& 0.335& 0.378& 0.375& 0.337& 0.312& 0.263& 0.237& 0.172\\\midrule
1.58M & 0.192& 0.316& 0.330& 0.366& 0.374& 0.337& 0.309& 0.275& 0.249& 0.192\\\midrule
Diff & \underline{-0.014}& 0.004& \underline{-0.005}& \underline{-0.012}& -0.001& 0.00& -0.003& \textbf{0.012}& \textbf{0.012}& \textbf{0.020}\\\midrule[1.5pt]
\multicolumn{11}{c}{QUAK-M (1.58M)} \\\midrule
100K & 0.210& 0.295& 0.342& 0.380& 0.393& 0.361& 0.339& 0.303& 0.269& 0.223\\\midrule
1.58M & 0.215& 0.329& 0.361& 0.398& 0.413& 0.381& 0.366& 0.340& 0.312& 0.265\\\midrule
Diff & \underline{0.005}& 0.034& \underline{0.019}& \underline{0.018}& 0.020& 0.020& 0.027& \textbf{0.037}& \textbf{0.043}& \textbf{0.042}\\\midrule[1.5pt]
\multicolumn{11}{c}{QUAK-M (6.58M)}\\\midrule
100K & 0.210& 0.295& 0.342& 0.380& 0.393& 0.361& 0.339& 0.303& 0.269& 0.223\\\midrule
6.58M & 0.184& 0.308& 0.352& 0.400& 0.399& 0.377& 0.364& 0.339& 0.298& 0.261\\\midrule
Diff & \underline{-0.026}& 0.013& \underline{0.010}& 0.020& \underline{0.006}& 0.016& 0.025& \textbf{0.036}& \textbf{0.029}& \textbf{0.038}\\\bottomrule[1.5pt]
\end{tabular}}\caption{\label{tb4:result_range}Target MCC performance difference (Diff) by TER range for Google test set. When we add data, we underline the three cases with the worst performance, and bold the three cases with the most performance improvement.}
\end{table*}

\paragraph{Performance of QUAK-M (6.58M)}
QUAK-M requires only monolingual corpus in the data building process. This allows data size expansion over other sub-QUAKs that utilize a parallel corpus. Exploiting these, we further extend the Wikipedia corpus by 5M, comprising a total of 6.58M. We gradually add data in 500K increments to check the performance fluctuation.

The experimental result is presented in Table \ref{tb4:result_mono}. The target MCC performance on the Google test set with 1.58M is lower by -0.016 at 3.58M(+2M) and -0.020 at 5.58M(+4M). The performance of Amazon, Microsoft, and Systran also degraded by -0.016, -0.017, -0.008 at 3.58M and -0.02, -0.022, and -0.012 at 5.58M compared to 1.58M, respectively. We observe that the overall QE model performance has deteriorated as more data is added.

We interpret this result in terms of data. QUAK-H, P, and the test sets are extracted from the Ai hub dataset. As observed in the previous statistics (Table \ref{tb:stat}), this resulted in a difference in terms of average TER in QUAK-M, which also contains Wikipedia. QUAK-M is mainly distributed in a range with a high TER score, while QUAK-H, QUAK-P, and the test set are included in relatively low scores. This indicates that the difference from the test set in terms of data distribution also affected the performance of QUAK-M.

\paragraph{Performance Comparison by TER Range}
In addition to the previous results, we divide the Google test set into units of 0.1 TER for more precise comparison. We then verify the changes in performance with the TER range. The experimental result is present in Table \ref{tb4:result_range}, from which we note that the performance on QUAK-M (6.58M) shows an overall improvement compared with those on QUAK-M (100K), and mainly improves between 0.7--1.0. The highest increase for QUAK-M (1.58M) is also seen between 0.7--1.0. As shown in Figure \ref{fig1:data_stat}, QUAK-M is mainly distributed in the high TER range. Although both QUAK-M (1.58M) and QUAK-M (6.58M) show performance gains over 100K, QUAK-M (6.58M) reports that the performance improvement is not significant at the relatively low TER. We analyze that this in turn, leads to performance degradation of the integrated score compared with 1.58M. This is supported by the fact that even in the TER range of 0.1--0.2, the performance fluctuation of QUAK-M (6.58M) is remarkably lower than that of QUAK-M (1.58M).

In QUAK-P, the amount of data is the lowest at 0.0--0.2 and 0.8--1.0. Therefore, when adding data, the performance variation also shows a lower increase compared with other scores in the range of 0.0--0.1 and 0.8--1.0. From the above results, we conclude that the amount of data can be a contributing factor for performance improvement.

\section{Conclusion}
We expose three drawbacks in terms of manual QE data construction: human labor and time cost, resulting in limited amount of data and limited language pairs. Taking this into account, we present QUAK, a synthetic Korean-English QE dataset for word-level QE. We automatically generated three sub-QUAKs with three strategies and quantitatively analyzed the trained QE models using them. First, QUAK-P is generated based on parallel corpus and induced the best performance among three sub-QUAKs. Along with QUAK-P, an increase in the data size of QUAK-M had a positive effect on performance gain. However, in further expansion using Wikipedia, the improvement in the low TER range was poor, so the overall performance fell. The QE model trained with QUAK-H did not show a steady performance gain. 

This dataset was built in a fully automated manner, eliminating human intervention while increasing reusability and scalability. The QUAK dataset generation process is language-agnostic if there is an MT model and corresponding corpus (monolingual or parallel). 
% In future work, we will endeavor to produce a multilingual-QUAK (mQUAK) for various language pairs based on this paper.

\section{Acknowledgements}
This work was supported by Institute of Information \& communications Technology Planning \& Evaluation(IITP) grant funded by the Korea government(MSIT) (No. 2020-0-00368, A Neural-Symbolic Model for Knowledge Acquisition and Inference Techniques), and supported by the MSIT, Korea, under the Information Technology Research Center(ITRC) support program(IITP-2022-2018-0-01405) supervised by the IITP, and supported by Basic Science Research Program through the National Research Foundation of Korea(NRF) funded by the Ministry of Education(NRF-2021R1A6A1A03045425).

\bibliography{anthology}
\bibliographystyle{acl_natbib}

%\appendix

%\section{Example Appendix}
%\label{sec:appendix}

\end{document}